\documentclass[twoside]{article}
\usepackage[T1]{fontenc}    
\usepackage{nicefrac}       
\usepackage{microtype}      
\usepackage{authblk}
\usepackage{latexsym}
\usepackage{amsmath, amsthm, geometry}
\usepackage{algorithm}  
\usepackage{graphicx}
\usepackage[round]{natbib}

\usepackage{url} 
\usepackage{float}
\usepackage{xcolor}
\usepackage{setspace} 
\usepackage{enumerate} 
\usepackage{mathtools}
\usepackage{mathrsfs}
\usepackage{booktabs}
\usepackage{multirow}
\usepackage{caption}
\usepackage{subcaption}
\usepackage{bm}
\usepackage{amssymb}
\usepackage{array}
\usepackage{algpseudocode}
\usepackage{hyperref}
\newtheorem{definition}{Definition}
\newtheorem{theorem}{Theorem}
\newtheorem{proposition}{Proposition}[section]

\newtheorem{remark}{Remark}[section]
\newtheorem{corollary}{Corollary}

\newtheorem*{lemma*}{Lemma}
\usepackage[accepted]{aistats2026}
\newcommand{\Input}{\item[\textbf{Input:}]}
\newcommand{\Output}{\item[\textbf{Output:}]}
\usepackage[normalem]{ulem}

\begin{document}

\twocolumn[

\aistatstitle{Differentially Private Linear Regression and Synthetic Data Generation with Statistical Guarantees}

\aistatsauthor{ Shurong Lin \And Aleksandra Slavkovi\'c \And  Deekshith Reddy Bhoomireddy }

\aistatsaddress{ The Pennsylvania State University } ]

\begin{abstract}
\vspace{-0.5em}
In the social sciences, small- to medium-scale datasets are common, and linear regression is canonical. In privacy-aware settings, much work has focused on differentially private (DP) linear regression, but mostly on point estimation with limited attention to uncertainty quantification. Meanwhile, synthetic data generation (SDG) is increasingly important for reproducibility studies, yet current DP linear regression methods do not readily support it. Mainstream DP-SDG approaches either are tailored to discrete or discretized data, making them less suitable for analyses involving continuous variables, or rely on deep learning models that require large datasets, limiting their use for the smaller-scale data typical in social science. We propose a method for linear regression with valid inference under Gaussian DP. It includes a bias-corrected estimator with asymptotic confidence intervals (CIs) and a general SDG procedure such that the corresponding regression on the synthetic data matches our DP linear regression procedure. Our approach is effective in small- to moderate-dimensional settings. Experiments show that our method (1) improves accuracy over existing methods for DP linear regression, (2) provides valid CIs, and (3) produces more reliable synthetic data for downstream statistical and machine learning tasks than current DP synthesizers.

\end{abstract}

\section{Introduction}
In the social, economic, and behavioral sciences, where small- to medium-scale datasets are common, linear regression~(LR) and subsequent statistical inference are widely used to address important scientific questions. Data from these contexts readily contain sensitive information. The  confidentiality protection methodology for sharing data and the results of statistical analyses has a long history drawing from many fields (e.g., \cite{hundepool2012statistical, slavkovseeman2023}). The modern methods  predominantly rely on differential privacy (DP) \citep{Dwork2006cal} 
to ensure rigorous privacy guarantees. Numerous methods for fitting LR under DP  have been proposed, including general approaches such as objective perturbation \citep{kifer12, Zhang2012Functional} and DP (stochastic) gradient descent (DP-SGD) \citep{Bassily2014PrivateER, Abadi2016}, as well as LR-specific techniques like sufficient statistics perturbation (SSP) and its variants \citep{DworkTT014, wang2018}. However, most methods focus on point estimation and provide statistical risk bounds but with limited support for uncertainty quantification. Valid statistical inference in LR settings under DP remains a challenge due to inadequate accounting for the noise added to satisfy the privacy guarantee.
\cite{Sheffet17} derived theoretical inference results for the SSP and Johnson--Lindenstrauss Transform (JLT) mechanisms, but some of these rely on the strong assumption of a Gaussian design matrix. Moreover, the SSP and JLT approaches exhibit substantially larger error than our more flexible method and other baselines (see Figure~\ref{fig:error}), and therefore offer limited practical utility.

Meanwhile, reproducibility and replicability are important concepts in trustworthy social science research \citep{nasem2019reproducibility, webb2026assessingutilitydifferentialprivacy}. Researchers often want to conduct replication studies to verify or build upon prior analyses.
In privacy-aware settings, however, typical DP methods only return model estimates, preventing others from revisiting or extending the analysis without access to the original data. Synthetic data generation (SDG) offers a possible solution. The basic idea was proposed in the early 1990s, but its broad adoption is still lacking~\citep{VANKESTEREN2024101049}. Furthermore, most methods for generating synthetic data under DP either rely on large datasets and complex models, such as deep learning--based approaches~\citep{Jordon2018PATEGANGS,Xie2018,xin2020private,xin2022federated}, or discretize continuous variables and produce discretized synthetic data, thereby sacrificing continuity~\citep{McKenna2022AIM,Zhang2021PrivSyn,Cai2021DataSyn}. These limitations restrict their applicability in small- to medium-scale settings, particularly when preserving continuity is essential. Moreover, the statistical implications of conducting downstream analyses, including LR, on such DP synthetic data remain largely unexplored, especially with respect to inferential validity.

To address these challenges, we propose a novel unified method for LR and SDG under Gaussian DP that provides valid statistical inference through an effective and practical binning-aggregation strategy. We use an existing DP binning method as a preprocessing step to obtain a DP partition of the covariate domain for aggregation.  Unlike approaches that rely on discretizing continuous variables, this step does not force the final synthetic data to lie on a discrete support or replace continuous variables by categorical values. The novelty lies in the binning-aggregation framework: by aggregating covariates and responses within bins, we reformulate LR as a weighted model, which supports valid statistical inference and provides a general procedure for SDG under DP. To our knowledge, this is the first work to reformulate DP-LR in this way, offering both inference guarantees and the ability to generate synthetic data within the same framework.

\vspace{-0.4em}
\paragraph{Main Contributions.}
(1) We propose a method for LR that satisfies Gaussian DP and mostly achieves the lowest estimation error among existing DP-LR algorithms, particularly on real datasets; 
see Algorithm \ref{alg:regression} and Theorem \ref{thm:priv}. Our method requires minimal tuning and runs significantly faster than computationally intensive approaches. (2)
We develop a DP statistical inference procedure based on the central limit theorem (CLT), analogous to the classical non-private regression, without requiring any assumptions on the covariate distribution. 
A CLT result for DP-LR has been missing from the literature, and our work provides the first such statement; see Theorem \ref{thm:normality}. (3)
We introduce a SDG mechanism that provides a general procedure beyond LR and supports replication studies at no additional privacy cost; see Algorithm \ref{alg:synthetic_data} and Theorem \ref{thm:equiv}.

\subsection{Related Work}
\label{sec:related}
\vspace{-0.3em}
Existing DP-LR methods include objective function perturbation \citep{kifer12, Zhang2012Functional}, DP stochastic gradient descent (DP-SGD) \citep{Bassily2014PrivateER, Abadi2016, cai2021}, and one posterior sampling (OPS) \citep{Dimitrakakis2014,NIPS2016_a7aeed74}. These approaches are general-purpose, but typically require careful hyperparameter tuning. \cite{wang2018} proposed modified versions of sufficient statistics perturbation (SSP) \citep{DworkTT014, Sheffet17} and OPS, namely AdaSSP and AdaOPS, respectively, by introducing adaptive regularization.

A few more recent works of interest have a relatively narrow focus.
\cite{Alabi2022Differentially} proposed a DP method exclusively for simple linear regression that outputs predictions only at $x = 0.25$ and $0.75$, assuming $x$ and $y$ are within $[0,1]$.
\cite{varshney22a} proposed theory for a one-pass mini-batch SGD method for sub-Gaussian data via adaptive clipping, while 
\cite{milionis22a} focused on LR with unbounded covariates but assumed the covariates are Gaussian; neither work includes numerical evaluations.
\cite{amin2023easy} proposed a bound-free method relying on a Propose-Test-Release check, which often fails when $n$ is small. Their evaluations focused on datasets with \(n \gtrsim 1000 \cdot d\), making the method unsuitable for the smaller datasets we are targeting.
\cite{dick2023better} proposed a method to improve prediction accuracy through covariate selection rather than estimating the regression coefficient in the specified model.

While most of these methods provide statistical risk bounds, statistical inference has received much less attention.
Unlike the non-private setting, where inference is well-established, DP-LR lacks broadly applicable methods for uncertainty quantification. 
Although empirical approaches such as bootstrapping can be used to construct confidence intervals, they often require generating many estimates, which in turn necessitates splitting the privacy budget across multiple runs, or incur additional privacy costs for estimating regression errors \citep{ferrando2022parametric}. Overall, analytical solutions for valid LR inference under DP remain limited. \cite{Sheffet17} studied inference for SSP and JLT mechanisms, but some results rely on Gaussian design assumptions. \cite{Lin2024Differentially} derived approximate variance formulas for DP-GD and SSP in the context of linked data, where LR appears as a special case.

\section{Preliminaries}
In this section, we review some preliminaries on DP and an existing DP binning algorithm that is used as a preprocessing step in our method. 

\vspace{-0.4em}
\subsection{Differential Privacy}

\vspace{-0.5em}
Two concepts central to DP are \textit{neighboring relations} and \textit{sensitivity}. 
Let $\mathcal{X}$ be some data space, and $D, D' \in \mathcal{X}^\mathbb{N}$ be two \textit{neighboring datasets}, where one is obtained from the other by adding or removing a single record. This relation is denoted by $D\sim D'$. We refer to it as \textit{remove-one/add-one} neighboring relation.
The sensitivity of a function is defined as follows.
\begin{definition}[Sensitivity]
Consider the problem of privately releasing a statistic $\theta(D)$ of the dataset $D$. 
The \textit{sensitivity} of $\theta$ is defined as
\[
\operatorname{sens}(\theta) = \sup_{D \sim D'} |\theta(D) - \theta(D')|.
\]
\end{definition}

\begin{definition}[$(\varepsilon,\delta)$-DP, \cite{DworkR14}]
Let $\varepsilon>0, \delta > 0$. An algorithm $A$ is  
$(\varepsilon,\delta)$-differentially private,
if for every pair of neighboring datasets $D\sim D'$, and any possible output set $S$,
\begin{equation}
    \mathbb{P}(A(D)\in S) \leq e^\varepsilon \cdot \mathbb{P}(A(D')\in S) + \delta.
\end{equation}
\end{definition}
This notion is referred to as \textit{approximate DP}. When $\delta = 0$, this notion is commonly denoted $\varepsilon$-DP and is called \textit{pure DP} \citep{Dwork2006cal}.

In our work, we adopt Gaussian DP, a variant with a better statistical interpretation and a tighter composition property.
Consider the hypothesis testing:
\[H_0: \text{the distribution is } P  \text{\;\;vs.\;}  H_1: \text{the distribution is } Q.\]
Let $\alpha_\phi$ and $\beta_\phi$ denote Type I and II errors for rejection rule $\phi$.
Let $T(P,Q)(\alpha):=\inf_{\phi}\{\beta_\phi: \alpha_\phi \leq \alpha \}.$

\begin{definition}[Gaussian DP, \citet{dong2022gaussian}]
Let $\mu>0$. An algorithm $A$ is $\mu$-Gaussian differentially private ($\mu$-GDP),
if for every neighboring $D\sim D'$ and any $\alpha\in(0,1)$
\begin{equation*}
    T(A(D), A(D'))(\alpha) \geq T(\mathcal{N}(0, 1), \mathcal{N}(\mu, 1))(\alpha).
\end{equation*}
\end{definition}

In other words, testing ``$H_0$: the underlying dataset is $D$'' versus ``$H_1$: the underlying dataset is $D'$\,'' is at least as hard as testing ``$H_0$: the distribution is $\mathcal{N}(0, 1)$'' versus ``$H_1$: the distribution is $\mathcal{N}(\mu, 1)$.''

Gaussian DP and approximate DP are precisely mutually convertible, and pure DP implies Gaussian DP.
We use the following conversion properties in our work.

\begin{proposition} [Conversion]
\label{prop:conversions}
Let $\Phi(\cdot)$ be the standard Gaussian cumulative distribution function.
\begin{enumerate}[(i)]
\item (Corollary 2.13, \cite{dong2022gaussian}) A mechanism is $\mu$-GDP if and only if it is 
$(\varepsilon, \delta(\varepsilon))$-DP for all $\varepsilon > 0$, where
\(
\delta(\varepsilon) = \Phi\left( -\frac{\varepsilon}{\mu} + \frac{\mu}{2} \right)
- e^{\varepsilon} \Phi\left( -\frac{\varepsilon}{\mu} - \frac{\mu}{2} \right).
\)
\item (Theorem 5.1, \cite{Liu2022IdentificationAA}) Any $\varepsilon$-DP algorithm is also $\mu$-GDP for\\
\(
\mu = -2 \, \Phi^{-1} \left( \frac{1}{1 + e^{\varepsilon}} \right) \leq \sqrt{\frac{\pi}{2}}.
\)
\end{enumerate}

\end{proposition}

Like the classic notion of pure and approximate DP, Gaussian DP has the following fundamental properties and the Gaussian Mechanism \citep{dong2022gaussian}.
\begin{proposition}[Composition]
\label{propo:comp}
The $n$-fold composition of $\mu_i$-GDP mechanisms is 
$$
\sqrt{\mu_1^2 + \mu_2^2 + \cdots + \mu_n^2}.
$$
\end{proposition}

\begin{proposition}[Post-processing]
\label{propo:post}
If an algorithm $A$ is $\mu$-GDP, then any post-processing function $f$, i.e., $f \circ A$, is also $\mu$-GDP.

\end{proposition}

\begin{proposition}[Gaussian mechanism]
\label{propo:GausMech}
Define the Gaussian mechanism applied to a statistic $\theta$ on dataset $D$ by
\[
A(D) = \theta(D) + \xi,
\]
where $\xi \sim \mathcal{N}(0, \operatorname{sens}(\theta)^2 / \mu^2)$.
Then $A(\cdot)$ satisfies $\mu$-GDP.
\end{proposition}

\subsection{A DP Binning Algorithm}

The proposed method involves creating bins. 
If the binning strategy is data-independent, the bin boundaries are public and do not reveal sensitive information. 
For instance, one can set fixed bin widths before accessing sensitive data. However, data-independent strategies often lack adaptability and are prone to the curse of dimensionality.
To address this, one may opt for a data-dependent binning method, such as recursive partitioning based on counts, which produces a more refined and representative histogram.
This approach, however, incurs additional privacy cost, as the binning must itself be performed in a DP manner. In our work, we use the PrivTree algorithm of \cite{Zhang2016PrivTreeAD} to output private bins without counts.

The PrivTree algorithm builds a hierarchical, tree-structured partitioning of the data domain by recursively splitting nodes based on privately perturbed, down-biased counts. For a node $v$ at depth $d=\mathrm{depth}(v)$ (root has depth 0) with count $c(v)$, a penalized score is defined by subtracting a fixed amount $\tau$ per level so deeper nodes need stronger evidence to split:
\[
b(v)=\max\{\,c(v)-d\,\tau,\;\theta-\tau\,\}.
\]
Add Laplace noise to protect privacy,
\[
\hat b(v)=b(v)+\text{Laplace noise},
\]
and node $v$ is split if $\hat{b}(v) > \theta$.
A full algorithm and exact noise calibration can be found in Appendix~\ref{app:privtree}.
The choice of $\tau$ is determined by the desired privacy level, while $\theta$ is a tunable hyperparameter. As discussed in \cite{Zhang2016PrivTreeAD}, the negative bias helps ensure that setting the threshold to $\theta = 0$ typically results in sufficiently large point counts in each node.

In our implementation, the root node corresponds to the initial bin. We pass in a non-sensitive $d$-dimensional region represented as the Cartesian product $\Pi_{i=1}^d (L_i, U_i)$, where $L_i$ and $U_i$ denote the lower and upper bounds of the $i$-th covariate, respectively. Each node is recursively split along its widest dimension.
We use the resulting leaf nodes (final bins) in our algorithm design in Section~\ref{sec:meth}. Since PrivTree satisfies $\epsilon$-DP and our method adopts $\mu$-GDP, we leverage Proposition~\ref{prop:conversions}~(ii) to convert the privacy guarantees.

\section{Methodology}
We present \textit{BinAgg}, a framework with three algorithms: (1) the fundamental binning--aggregation step, (2) DP linear regression, and (3) DP synthetic data generation. A corresponding Python package is available on \href{https://github.com/Shuronglin/BinAgg}{GitHub}.
\label{sec:meth}
\subsection{Aggregated Linear Model}
Let \( X \) denote the \( n \times d \) matrix of covariates and \( \bm{y} \) be the $n$-dimensional response vector. The classic linear model is given by
\begin{equation}
\label{original_model}
    \bm y = X \bm{\beta} + \bm{e}, \quad  \bm{e} \sim \mathcal{N}(0, \sigma^2 \bm{I}_n).
\end{equation}
We propose an alternative formulation given the observations of $X$ are partitioned into $K$ bins, represented as $\{(\mathcal{B}_k, c_k)\}_{k=1}^{K}$, where $\mathcal{B}_k$ denotes the $k$th bin and $c_k$ is the number of observations in that bin. Let $j$ be the index over data points.
We aggregate the observations in each bin by defining
\[\bm s_k = \sum_{\bm x_j \in \mathcal{B}_k} \bm x_j, \quad t_k = \sum_{\bm x_j \in \mathcal{B}_k} y_j, \quad \eta_k = \sum_{\bm x_j \in \mathcal{B}_k} e_j.\]
In matrix form, we let
\(\displaystyle
S = (\bm{s}_1^\top, \bm{s}_2^\top, \dots, \bm{s}_K^\top)^\top, \quad
\bm t = (t_1, t_2, \dots, t_K)^\top, \quad
\bm \eta = (\eta_1, \eta_2, \dots, \eta_K)^\top.
\)
This aggregation leads to a weighted linear model:
\begin{equation}
\label{sum_model}
    \bm t = S \bm{\beta} + \bm \eta,
\end{equation}
where $\bm \eta \sim \mathcal{N}(0, \sigma^2 C)$ with $C = \operatorname{diag}(c_1, \dots, c_K)$. Let $W = C^{-1}$. An unbiased and consistent estimator of $\bm{\beta}$ is given by the weighted least squares (WLS) estimator:
\begin{equation}
\label{eq:hat_beta}
    \hat{\bm{\beta}} = (S^\top W S)^{-1} S^\top W \bm t.
\end{equation}

\begin{remark}
    Model (\ref{sum_model}) is equivalent to the averaged (weighted) model defined as 
    $\bm{\bar {y}} = \bar X \bm{\beta} + \bm{\bar {e}}$,
    where $\bar X  \stackrel{def}{=} C^{-1}S$, 
    $\bm{\bar {y}} \stackrel{def}{=}  C^{-1}\bm t$, and 
    $\bm{\bar {e}} \stackrel{def}{=} C^{-1} \bm \eta \sim \mathcal{N}(0, \sigma^2 C^{-1})$. This model yields the same WLS estimator $\hat{\bm{\beta}} = (\bar X ^\top C \bar X )^{-1} \bar X 
    ^\top C \bm{\bar {y}} = (S^\top W S)^{-1} S^\top W \bm t$ as in (\ref{eq:hat_beta}).
    In this work, we proceed with Model (\ref{sum_model}) for DP algorithm design.
\end{remark}

Our approach that ensures DP for the original data $(X, \bm y)$ consists of two major steps:
(1) apply a DP algorithm to determine the bin structure $\{\mathcal{B}_k\}_{k=1}^{K}$, with PrivTree being one option, and (2) add noise to the aggregated statistics, specifically to $\bm t$, $S$, and the count matrix $C$. Then a DP aggregated model is
\begin{equation}
\label{pro_model}
    \widetilde{\bm t} = \widetilde{S} \bm{\beta} + \tilde{\bm \eta},
\end{equation}
where $\tilde{\bm \eta} \sim \mathcal{N}(0, \sigma^2 \widetilde{C})$. Let $\widetilde{W} = \widetilde{C}^{-1}$. A naive estimator is given by
\begin{equation*}
    \tilde{\bm{\beta}}_{\text{naive}} = (\widetilde{S}^\top \widetilde{W} \widetilde{S})^{-1} \widetilde{S}^\top \widetilde{W} \widetilde{\bm{t}}.
\end{equation*}
However, it does not account for the extra uncertainty introduced by injected noise.
Instead, we propose a debiased estimator (see Theorem~\ref{thm:normality}): 
\begin{equation*}
    \tilde{\bm{\beta}} = (\widetilde{S}^\top \widetilde{W} \widetilde{S} - \widetilde{D})^{-1} \widetilde{S}^\top \widetilde{W} \widetilde{\bm{t}},
\end{equation*}
where $\widetilde{D}$ is a private bias-correction matrix, as specified in Algorithm~\ref{alg:regression}.

Although the aggregated model involves $K$ bins as the effective sample size rather than $n$ individual observations, this does not necessarily imply a substantial loss of statistical efficiency. After accounting for the weights, the variance of each bin-level summary shrinks with its count $c_k$. 
In other words, aggregation reduces the effective size but simultaneously reduces variability of the effective random error. Moreover, adaptive DP partitioning (e.g., PrivTree) mitigates extreme or highly unbalanced binning structure by allocating finer partitions in dense regions and coarser ones elsewhere, helping preserve utility in practice.

\subsection{DP Binning-Aggregation Algorithms}
Motivated by Section~3.1, we propose the binning–aggregation (BinAgg) framework, as captured by Algorithm~\ref{alg:binagg-core}. The novelty of BinAgg lies in converting raw data into a set of DP bin-level summaries sufficient for both LR and SDG. Given a DP partition of the covariate space, Algorithm~\ref{alg:binagg-core} aggregates records within each bin to form counts and per-bin sums of covariates and responses. It releases privatized bins and counts under task-specific privacy budgets, while the per-bin sums of covariates and responses are privatized in later algorithms.
The partition requires a prespecified, public domain $\mathcal{X}$ for $X$,
provided by the analyst (e.g., based on survey design or known variable ranges). If such bounds are not naturally available or are too conservative, they may be obtained via standard clipping based on domain knowledge or privately estimated using a small portion of the privacy budget. The same assumption applies to the response variable $y$, for which a public bound is also specified.
By injecting privacy noise to bin-level sums, rather than to per-record quantities or sufficient statistics, the framework (i) preserves the joint $(X,\bm y)$ structure and (ii) reduces effective sensitivity: each coordinate is confined to its bin range, and bins with small (privatized) counts can be discarded. The contribution is a novel coupling of DP binning with aggregation to yield a weighted linear model for valid inference and a general SDG mechanism simultaneously.

\begin{algorithm}[h]
\caption{\small DP BinAgg Preparation}
\label{alg:binagg-core}
\begin{algorithmic}[1]
\Input Dataset $(X, \bm{y})$, domain for $X$, privacy budgets for binning and counts: $\mu_{\text{bin}}$ and $\mu_c$.
\State Create a list of $\mu_{\text{bin}}$-GDP bins for $X$ (e.g., via PrivTree) : $\{\mathcal{B}_k\}_{k=1}^{K}$ where $\mathcal{B}_k = \Pi_{i=1}^d(L_{ki}, U_{ki})$
\State For each bin $\mathcal{B}_k$, compute:
    $ c_k = \sum_{\bm{x}_j \in \mathcal{B}_k} 1$
\For{$k = 1$ to $K$}
    \State Privatize count:
    $$\tilde{c}_k = \mathrm{round}(c_k + \xi^c), \quad \xi^c \sim \mathcal{N}(0, 1/\mu_c^2)
   $$
    \If{$\tilde{c}_k < 2$}
        \State Discard bin $\mathcal{B}_k$.
    \EndIf
\EndFor
\State Reset $K$ to be the number of bins after discarding.

\State For each bin $\mathcal{B}_k$, compute:
     $$\bm{s}_k = \sum_{\bm{x}_j \in \mathcal{B}_k} \bm{x}_j, \quad t_k = \sum_{x_j \in \mathcal{B}_k} y_j$$
\State Compute sensitivity vector $\bm{\Delta}_{k} = (\Delta_{k1}, \dots, \Delta_{kd})^\top$, where $\Delta_{ki} = \max(|L_{ki}|, |U_{ki}|)$
\Output Privatized bins and counts $\{(\mathcal{B}_k,\tilde c_k)\}_{k=1}^K$; bin-wise aggregates $\{(\bm s_k,t_k)\}_{k=1}^K$ (to be privatized) with sensitivity vectors $\{\bm\Delta_k\}_{k=1}^K$ for $\bm s_k$.
\end{algorithmic}
\end{algorithm}

\vspace{-1em}
\paragraph{Privacy Model.}
Throughout this paper, we adopt the unbounded notion of DP, under which neighboring datasets differ by adding or removing a single record (i.e., the remove-one/add-one relation), whereas bounded DP uses the replace-one neighboring relation. The unbounded notion is common in the DP synthetic data literature and is often preferred when either definition is acceptable, since it yields lower sensitivity than bounded DP under the same privacy parameters and therefore requires less noise~\citep{McKenna2022AIM}. That said, the two notions are qualitatively different, as they are based on different neighboring relations, so the same privacy parameters do not have the same interpretation. Our framework can also be adapted to bounded DP by recalibrating the sensitivities and the corresponding Gaussian noise under the replace-one neighboring relation.

\paragraph{Sensitivity of $\bm x$ and $y$.}
Under the remove-one/add-one neighboring relation, a differing record affects only one bin. Let $i$ index the covariate dimensions in the $k$th bin, denoted by $\mathcal{B}_k=\prod_{i=1}^d (L_{ki},U_{ki})$. For the per-bin sum of covariates, the coordinate-wise sensitivity is
$
\Delta_{ki}=\max\{|L_{ki}|,|U_{ki}|\},
$ and for the per-bin sum of responses it is given by the bound on the response variable, denoted by $B_y$. We collect $\bm \Delta_k=(\Delta_{k1},\ldots,\Delta_{kd})^\top$ to calibrate the Gaussian mechanisms in Algorithms \ref{alg:regression} and \ref{alg:synthetic_data}.

\begin{algorithm}[h] \caption{\small DP BinAgg for Linear Regression} \label{alg:regression} \begin{algorithmic}[1] \Input Output from Algorithm~\ref{alg:binagg-core}, response variable bound $B_y$, privacy budgets for covariates and response: $\mu_s$ and $\mu_t$  
\For{$k = 1$ to $K$} 
\State Privatize $\bm s_k$, and $t_k$: 
\begin{align*} 
\tilde{\bm s}_k &= \bm s_k + \bm \xi^s, \quad \bm \xi^s \sim \mathcal{N}(\bm 0, \bm \Delta_{k}^2 /\mu^2_s), \label{eq:tilde_ck} \\ \tilde{t}_k &= t_k + \xi^y, \quad \xi^y \sim \mathcal{N}(0, B^2_y /\mu^2_t). 
\end{align*} 
\EndFor 
\State Let 
\begin{align*}
    \widetilde{W} & = \operatorname{diag}(\tilde{w}_1, \tilde{w}_2, \dots, \tilde{w}_K), \quad \tilde w_k = 1/\tilde c_k\\
     \widetilde{S} & = (\tilde{\bm{s}}_1^\top, \tilde{\bm{s}}_2^\top, \dots, \tilde{\bm{s}}_K^\top)^\top \\
     \widetilde{\bm t} & = (\tilde{t}_1, \tilde{t}_2, \dots, \tilde{t}_K)^\top,
\end{align*}
\State Calculate matrix $\widetilde{D} = \frac{1}{K} \sum_{k=1}^K\tilde{w}_k D_k$ where \(D_k = \text{diag}(\bm \Delta_{k}^2 /\mu^2_s)\). 
\Output Private estimator $$\tilde{\bm\beta}=(\widetilde{S}^\top\widetilde{W}\widetilde{S}-\widetilde{D})^{-1}\widetilde{S}^\top\widetilde{W}\,\widetilde{\bm t}$$ and a DP-CI for each coordinate $j$,
\[
\tilde{\beta}_{j} \pm z_{\alpha/2}\,\mathrm{se}(\tilde{\beta}_{j}),
\]
where $\mathrm{se}(\tilde{\beta}_{j})$ is defined in Section~\ref{sec:ci} and $z_{\alpha/2}$ is the $(1-\alpha/2)$ quantile of $\mathcal{N}(0,1)$.

\end{algorithmic} 
\end{algorithm}

This BinAgg framework instantiates two procedures: (1) LR in Algorithm~\ref{alg:regression}, which implements the weighted model induced by bin-level aggregation and provides valid CIs as established in Theorem~\ref{thm:normality}; and (2) SDG in Algorithm~\ref{alg:synthetic_data}, which generates samples directly from the same bin-level summaries to support reproducibility and broader downstream analyses.
These procedures target different use cases: Algorithm~\ref{alg:regression} performs DP-LR and provides DP-CIs, whereas Algorithm~\ref{alg:synthetic_data} produces a reusable DP synthetic dataset supporting residual analysis, visualization, and fitting alternative models. 
Algorithm~\ref{alg:synthetic_data} is more general and subsumes Algorithm~\ref{alg:regression}, since applying the weighted/aggregated linear model to the synthetic data yields an estimator that matches Algorithm~\ref{alg:regression} in distribution (see Corollary~\ref{thm:equiv}). 
This consistency is essential for reproducible scientific research. It allows for synthetic data sharing and  valid LR inference without discrepancies or additional privacy cost.
We remark that applying the usual unweighted linear regression directly to the synthetic data does not yield the debiased estimator or the associated CIs provided by our method. Practitioners who want both synthetic data and the debiased estimator may embed Algorithm~\ref{alg:regression} within Algorithm~\ref{alg:synthetic_data}; we provide this option in our software package.

\begin{algorithm}[ht]
\caption{\small DP BinAgg for Synthetic Data}
\label{alg:synthetic_data}

\begin{algorithmic}[1]
\Input Output from Algorithm~\ref{alg:binagg-core}, response variable bound $B_y$, privacy budgets for covariates and response: $\mu_s$ and $\mu_t$. 

\For{$k = 1$ to $K$}
    \For{$i = 1$ to $\tilde{c}_k$}
            \State Sample synthetic covariates:
           $$
                \tilde{\bm{x}}^{(k,i)} = \frac{\bm{s}_k + \bm{\xi}^x}{\tilde{c}_k}, \quad \bm{\xi}^x \sim \mathcal{N}(\bm{0}, \tilde{c}_k \bm{\Delta}^2_{k} / \mu_s^2),
          $$
            \State Sample synthetic response:
         $$
                \tilde{y}^{(k,i)} = \frac{t_k + \xi^y}{\tilde{c}_k}, \quad \xi^y \sim \mathcal{N}(0, \tilde{c}_k B_y^2 / \mu_t^2).
         $$
        \EndFor
\EndFor
\Output A DP synthetic dataset: $$\mathcal{D}_{\text{syn}} = \left\{ (\tilde{\bm{x}}^{(k,i)}, \tilde{y}^{(k,i)}) \mid k = 1,\dots,K; \, i = 1,\dots,\tilde{c}_k \right\}.$$
\end{algorithmic}
\end{algorithm}

\begin{corollary}[Equivalence of Two Algorithms]
 \label{thm:equiv}
 In Algorithm \ref{alg:synthetic_data}, aggregate the synthetic data points in each bin by letting 
 $\tilde{\bm s}'_k = \sum_{i=1}^{\tilde c_k} \tilde{\bm{x}}^{(k,i)} $ and $\tilde{t}'_k = \sum_{i=1}^{\tilde c_k} \tilde{y}^{(k,i)}$.
 Then, 
  \[\tilde{\bm s}'_k \stackrel{d}{=} \tilde{\bm s}_k \sim \mathcal{N}(\bm s_k, \bm{\Delta}^2_{k} / \mu_s^2), \quad \tilde{t}'_k \stackrel{d}{=} \tilde{t}_k  \sim \mathcal{N}(t_k, B_y^2 / \mu_t^2),\]
 where $\tilde{\bm s}_k$ and $ \tilde{t}_k $ are defined in Algorithm \ref{alg:regression}.
\end{corollary}

A natural alternative for SDG is to post-process a private regression fit (e.g., $\bm y=X\bm\beta^{\mathrm{priv}}$ or $\bm y=X\bm\beta^{\mathrm{priv}}+\bm e$), but this either collapses variability onto a hyperplane or requires a private estimate of the residual variance, which demands extra privacy budget and sensitivity analysis. In contrast, Algorithm~\ref{alg:synthetic_data} generates samples by using the DP bin-level summaries, thereby preserving variability while remaining consistent with the regression model, without additional privacy cost.

\vspace{-1em}
\paragraph{Remark on Synthetic Sample Size.}
Under unbounded DP, neighboring datasets differ in size by one record, so the sample size could be protected. Accordingly, Algorithm~\ref{alg:synthetic_data} does not require the synthetic dataset to have exactly the same size as the original data. Instead, its size is determined by the privatized counts, namely,
$
\tilde n = \sum_{k=1}^K \tilde c_k,
$
after discarding bins with $\tilde c_k < 2$. Thus, the released synthetic data do not reveal the original sample size $n$. In our implementation, the threshold $\tilde c_k < 2$ typically reduces the synthetic sample size only slightly.

If a synthetic dataset of exact size $m$ is desired (for example, $m=n$ when $n$ is treated as public information), one may post-process the noisy counts by rescaling
$
\tilde c_k \leftarrow \tilde c_k \cdot \frac{m}{\tilde n}
$
and applying an integer apportionment rule such as the largest remainder (Hamilton) method or randomized rounding to obtain integer counts summing to $m$, with no additional privacy cost. We provide this option in our software package.

\paragraph{Remark on Binning Dependence.} 
While BinAgg requires a binned structure (i.e., the bin boundaries) as its foundation, it is not inherently tied to the use of any particular binning algorithm. This flexibility allows practitioners to tailor the binning procedure to their data characteristics and privacy constraints, while still retaining the theoretical guarantees of our method. We adopt PrivTree in our implementation because it is a practical, data-dependent binning method that adapts to the data density and has been successfully combined with DP-SDG in prior evaluation studies \citep{tao2022benchmarking}. However, other binning approaches, either data-independent (e.g., uniform partitioning) or alternative data-dependent methods, can also be used in conjunction with BinAgg, provided that the bin boundaries are released in a DP manner. Importantly, during the binning step, BinAgg allocates privacy budget only for constructing the bin structure, not for any additional statistics that some binning algorithms may output.

\section{Theoretical Results}
\label{sec:theory}
This section presents two-fold theoretical results that capture the privacy-utility tradeoff: (1) privacy guarantees, and (2) statistical inference guarantees. 
All proofs are provided in Appendix~\ref{app:d}.

\subsection{Privacy Guarantees}
\begin{theorem}[Gaussian DP Guarantees]
\label{thm:priv}
Algorithms \ref{alg:regression} and \ref{alg:synthetic_data} satisfy
$\sqrt{\mu_{\text{bin}}^2 + \mu_s^2 + \mu_t^2 + \mu_c^2}$-GDP. 
\end{theorem}
The GDP guarantee follows directly from the design of our mechanisms and the composition properties of DP. Specifically, the binning step employs the PrivTree algorithm, which in our implementation we show ensures $\mu_{\text{bin}}$-GDP. The remaining components, including the sum of covariates $\bm x$, the sum of response variable $y$, and the bin count, are each released using Gaussian mechanisms calibrated to privacy budgets $\mu_s$, $\mu_t$, and $\mu_c$, respectively. By composition, their privacy losses combine, yielding the overall guarantee of 
$\sqrt{\mu_{\text{bin}}^2 + \mu_s^2 + \mu_t^2 + \mu_c^2}$-GDP.

Algorithm~\ref{alg:regression} and Algorithm~\ref{alg:synthetic_data} both inherit the same overall privacy bound, and one may choose one or the other depending on the use case and analysis goals. In particular, by the post-processing property of DP, any subsequent analysis, including LR, performed on the synthetic data does not incur additional privacy cost.

\subsection{Statistical Inference}
\label{sec:ci}
In this section, we establish the asymptotic normality of the proposed private estimator and use it to construct asymptotic CIs, accounting for noise added for privacy protection.

For the non-private weighted linear model in (\ref{sum_model}), the classical theory gives
\[\Sigma^{-1/2}(\hat{\bm\beta} - \bm\beta) \xrightarrow{d} \mathcal{N}\left(0, \bm I_d\right), \quad  \Sigma = \sigma^2(S^\top W S)^{-1}.\]
In the private model (\ref{pro_model}), a naive plug-in estimator takes the form \(\widetilde\Sigma_{\text{naive}} = \sigma^2(\widetilde{S}^\top \widetilde{W} \widetilde{S})^{-1}\). However, the naive covariance estimator does not account for DP-induced uncertainty properly, leading to undercoverage of the resulting CI; see Table \ref{tab:CI}.
Instead, we provide the following theoretical result for the proposed DP bias-corrected estimator.

\begin{theorem}[Asymptotic Normality]
\label{thm:normality}
As $K\to\infty$ with $n\to\infty$, assume there exists a constant $c_0>0$
such that $\min_{1\le k\le K} c_k \ge c_0$,
and that the injected DP noises have finite variances. Define
\[
\widetilde M := \frac{1}{K}\Big(\widetilde{S}^\top \widetilde{W}\widetilde{S}\Big)-\widetilde D,
\]
and assume $\widetilde M \xrightarrow{p} M$ for some finite, nonsingular matrix $M$.
The bias-corrected estimator
\[
\tilde{\bm \beta}
:= (\widetilde{S}^\top \widetilde{W} \widetilde{S} - \widetilde{D})^{-1}
\widetilde{S}^\top \widetilde{W}\widetilde{{\bm{t}}}
\]
satisfies
\[
\sqrt{K}\,(\tilde{\bm\beta}-\bm\beta)\ \xrightarrow{d}\
\mathcal{N}\!\left(\bm 0,\; M^{-1} H M^{-1}\right),
\]
where
\begin{align*}
    H :& = \lim_{K\to\infty}\frac{1}{K}\sum_{k=1}^K \operatorname{Var}\!\big(\bm Q_k(\bm\beta)\big),
\\
\bm Q_k(\bm \beta)
:&= \tilde{\bm s}_k \tilde w_k\big(\tilde t_k-\tilde{\bm s}_k^\top \bm \beta\big)
   +\tilde w_k D_k\,\bm \beta.
\end{align*}

Moreover, a consistent and private estimator of $\operatorname{Var}(\tilde{\bm\beta})$ is given by
\[
\widetilde\Sigma
=\frac{1}{K}\,\widetilde M^{-1}\widetilde H\,\widetilde M^{-1},
\]
where
\begin{align*}
\widetilde H
& =: \frac{1}{K-d}\sum_{k=1}^K \widetilde{\bm Q}_k\widetilde{\bm Q}_k^\top,
\\
\widetilde{\bm Q}_k
& =:\tilde{\bm s}_k \tilde w_k(\tilde t_k-\tilde{\bm s}_k^\top \tilde{\bm\beta})
  +\tilde w_k D_k\,\tilde{\bm\beta}.
\end{align*}
Therefore, an asymptotic $(1-\alpha)$ confidence interval for $\beta_j$ is
\[
\tilde\beta_{j}\ \pm\ z_{\alpha/2}\,\sqrt{[\widetilde\Sigma]_{jj}},\qquad j=1,\dots,d.
\]
\end{theorem}

The asymptotic normality follows from the central limit theorem, as in the non-private LR setting. The proposed CI is differentially private because it is constructed entirely from privatized quantities. It also accounts for the uncertainty introduced by the injected DP noise, as reflected in the construction of the covariance estimator $\widetilde{\Sigma}$.

\section{Experiments}
In this section, we compare our algorithms with existing DP-LR and DP-SDG methods, and assess the validity of the private CIs from Theorem~\ref{thm:normality}.

\subsection{Simulation Studies}

We compare our Algorithm~\ref{alg:regression} to two popular algorithms: DP-(S)GD and AdaSSP (the only two methods that are used for comparison in~\cite{amin2023easy}).
Given our focus on small-scale datasets, we use the variant DP-GD for its more stable performance without the need to tune batch size.
In addition, we also compare it to SSP~\citep{DworkTT014, Sheffet17} and JLT~\citep{Sheffet17} that also give CIs.
The non-private OLS estimator is displayed for benchmarking purposes.

Covariates are drawn from $\text{Uniform}([0,1]^d)$, and the true coefficients are drawn from $\text{Uniform}([1,2]^d)$. The regression error scale is set to $\sigma = 1$. Details on the hyperparameter settings are provided in Appendix~\ref{app:a}.

\begin{figure}[ht]
  \centering
    \includegraphics[width=0.49\textwidth]{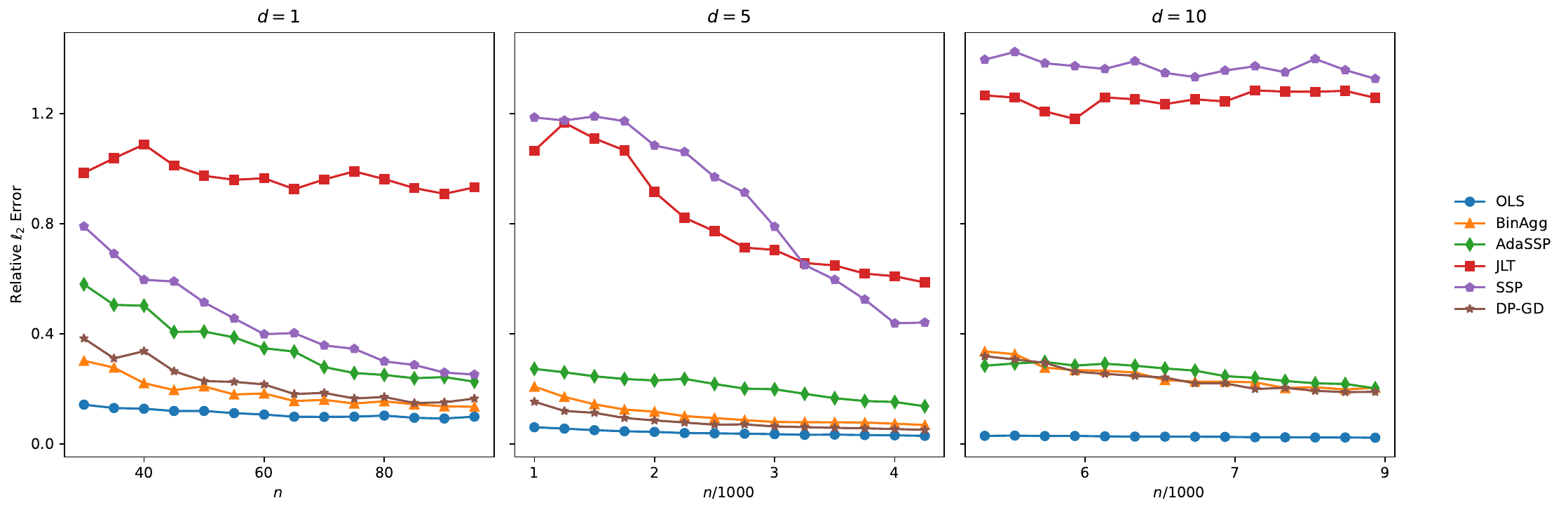}
  \caption{\small Coefficient estimation error across different $(n, d)$ over 100 repetitions with $\mu=1$}
  \label{fig:error}
\end{figure}

We evaluate five methods under three settings.
Figure~\ref{fig:error} shows the estimation error across varying sample sizes in three settings with dimensions $d = 1, 5, 10$.
The $y$-axis displays the averaged relative $\ell_2$ error over 100 repetitions, defined as
\( \|\tilde{\bm\beta}-\bm\beta\|_2/\|\bm\beta\|_2 \) where $\bm\beta$ denotes the true coefficients and $\tilde{\bm\beta}$ is any estimator.
Precise conversion in Proposition~\ref{prop:conversions} (i) is used for other $(\varepsilon, \delta)$-DP methods by setting $\delta = 1/n^{1.1}$.
The performance of the five algorithms falls into two tiers: SSP and JLT perform poorly, while the other three methods exhibit better accuracy.  
Among them, BinAgg and DP-GD perform comparably well overall, with AdaSSP slightly worse.  
Notably, BinAgg outperforms all others on the smallest-scale datasets with $d=1$.
At first glance, DP-GD may appear to perform slightly better in some settings (e.g., 
$d=5$). However, achieving this typically requires extensive hyperparameter tuning, which is \textit{computationally expensive} and can implicitly \textit{leak privacy} through the tuned parameters. In practice, identifying an effective tuning grid is harder than in simulation. By contrast, in our real-data evaluations (Section~\ref{subsec:real_data_lr}), DP-SGD mostly underperforms BinAgg. Moreover, neither DP-GD nor AdaSSP provide CIs.

\begin{figure}[ht]
  \centering
  \includegraphics[width=0.5\textwidth]{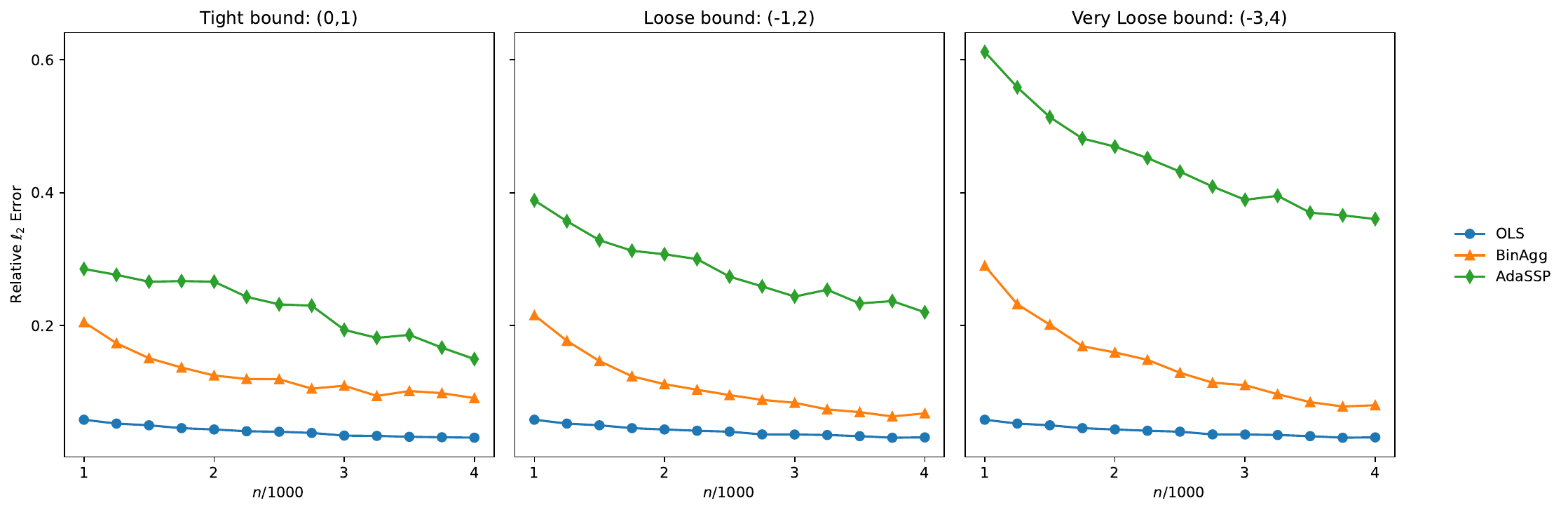}
  \caption{\small Coefficient estimation error using different covariate bounds with $d=5$ and $\mu=1$.}
\label{fig:loosebounds}
\end{figure}

In addition, BinAgg is less sensitive to loose covariate bounds than AdaSSP. Figure~\ref{fig:loosebounds} shows the error under varying bounds for $d = 5$. In AdaSSP, the covariate bounds directly determine sensitivity and thus the noise level. In contrast, BinAgg uses the bounds only to initialize bin boundaries, and bins with low noisy counts are discarded, mitigating the effect of loose initial bounds. More aggressive filtering of such low-count bins can be achieved by increasing the threshold.

We also evaluate DP-CIs based on Theorem \ref{thm:normality}, with results shown in Table \ref{tab:CI}; SD stands for standard deviation. The naive CI discussed in Section \ref{sec:theory} is included to illustrate that ignoring extra uncertainty due to DP underestimates the variance of the estimator, and the resulting CI significantly undercovers. Given that the JLT and SSP methods consistently give significantly poor performance, we do not include their CIs. Although they support uncertainty quantification, the estimation error is too large to be useful in practice. 
From Table \ref{tab:CI}, our theoretical standard deviations are very close to the empirical ones, and the empirical coverage is around the nominal level of 95\%.

\begin{table}[h!]
  \caption{\small BinAgg: Gaussian DP 95\% CI over $2000$ repetitions with $d=5$, $n=1000$, $\mu=1$.}
  \label{tab:CI}
  \centering
  \resizebox{0.5\textwidth}{!}{%
  \begin{tabular}{lcccccc}
    \toprule
    Avg. bias & Empirical SD & Avg. theor. SD & Naive theor. SD & Coverage & Naive coverage \\
    \midrule
    -0.012 & 0.252 & 0.255 & 0.113 & 0.953 & 0.637 \\
    -0.008 & 0.262 & 0.268 & 0.117 & 0.950 & 0.654 \\
    -0.002 & 0.271 & 0.271 & 0.119 & 0.947 & 0.637 \\
    -0.004 & 0.298 & 0.307 & 0.129 & 0.947 & 0.637 \\
     0.004 & 0.503 & 0.521 & 0.194 & 0.957 & 0.597 \\
    \bottomrule
  \end{tabular}%
  }
\end{table}

\subsection{Real Data Applications}

\subsubsection{Linear Regression}
\label{subsec:real_data_lr}
We apply Algorithm~\ref{alg:regression} to several small- to mid-scale real datasets accessible in the UCI Machine Learning Repository, covering application domains across various scientific areas. The data vary in size from $n=182$ to $n=21{,}263$ and in dimensionality from $d=4$ to $d=81$.
For brevity, we denote these datasets as D1–D9 throughout the text; additional details and references are provided in Appendix~\ref{app:b}.
Given the significantly worse performance of SSP and JLT, we focus our comparison on three methods: BinAgg, AdaSSP, and DP-GD, using non-private OLS as the benchmark.
To evaluate prediction accuracy, in Table~\ref{tab:lr_rmse} we report the relative mean squared error (MSE), defined as
$\|\tilde{\bm y} - \bm y\|_2^2/ \|\bm y\|_2^2,
$
where $\tilde{\bm y}$ denotes the predicted values. All methods on each dataset satisfy $\mu$-GDP with $\mu=1$.

\begin{table}[h]
\centering
\caption{\small Relative MSE of prediction across datasets over 100 repetitions. 
Lowest error per dataset in \textbf{bold}; second lowest \uline{underlined}. All DP methods satisfy $\mu$-GDP with $\mu=1$.}
\label{tab:lr_rmse}
 \resizebox{0.47\textwidth}{!}{%
\begin{tabular}{cccccc}
\toprule
Dataset & Size $(n, d)$ & OLS & BinAgg & AdaSSP & DP-GD \\
\midrule
D1 & (182, 4)    & 0.038 & \textbf{0.095} & 0.690 & 0.677 \\
D2 & (345, 6)    & 0.084 & \uline{0.151} & 0.229 & \textbf{0.102} \\
D3 & (2043, 8)   & 0.023 & \textbf{0.035} & 0.669 & 0.693 \\
D4 & (4177, 10)  & 0.044 & \textbf{0.059} & 0.082 & 0.059 \\
D5 & (5875, 21)  & 0.011 & \textbf{0.016} & 0.080 & 0.062 \\
D6 & (6497, 12)  & 0.016 & \textbf{0.022} & 0.203 & 0.120 \\
D7 & (9357, 12)  & 0.441 & \textbf{0.463} & 0.682 & 0.852 \\
D8 & (19735, 27) & 0.438 & \uline{0.507} & \textbf{0.500} & 0.546 \\
D9 & (21263, 81) & 0.131 & \textbf{0.203} & 0.429 & 0.420 \\
\bottomrule
\end{tabular}
}
\end{table}

For DP-GD, we use a grid search setup similar to \citet{amin2023easy}. For all other applicable methods, we use common non-private bounds computed from the observed dataset. These bounds are used solely for controlled comparison, following prior implementations that use non-private dataset-specific bounds or normalization \citep{amin2023easy,wang2018}. Further details and discussion are given in Appendix~\ref{app:a}. In practice, when public bounds are unavailable, one may instead use clipping based on domain knowledge or privately estimate bounds using a small privacy budget. While conservative bounds can reduce utility, BinAgg demonstrates greater robustness to loose bounds than AdaSSP; see Figure~\ref{fig:loosebounds}.

Table~\ref{tab:lr_rmse} shows that BinAgg outperforms the other two methods on most datasets, and at least achieves the second-lowest relative MSE. This demonstrates that BinAgg is \textit{both effective and robust in practice}, with its performance advantage more pronounced than in the controlled simulation settings. In contrast, AdaSSP performs the worst overall. Although DP-GD occasionally approaches BinAgg, it does not surpass it on most datasets. Moreover, DP-GD’s performance is highly sensitive to hyperparameter tuning. Achieving competitive results requires an extensive search, which comes at a substantially higher computational cost.

\subsubsection{Synthetic Data Generation}

Our private synthetic data supports both scientific reproducibility with privacy guarantees and broader downstream tasks.
To further assess downstream utility, we evaluate performance on additional widely used machine learning models for regression.
In this setting, we compare Algorithm~\ref{alg:synthetic_data} against five DP-SDG approaches: AIM~\citep{amin2023easy}, BinAgg, DP-GAN~\citep{Xie2018}, PATE-GAN~\citep{Jordon2018PATEGANGS}, and their enhanced variants DP-CTGAN and PATE-CTGAN~\citep{Xu2019CTGAN}. Implementation of all five algorithms is available in the open-source SmartNoise library~\citep{SmartNoise2021}.
AIM is a marginal-based method that requires discretizing the data, with the synthetic data inheriting the discretized structure. It has been shown to be the strongest marginal-based baseline~\citep{chenbenchmarking}. 
DP-GAN and PATE-GAN, together with their CTGAN extensions, represent approaches based on complex generative adversarial networks (GANs).

For each method, we generate 10 synthetic datasets and then train four regression models: XGBoost, Random Forest, Support Vector Regression (SVR), and Multilayer Perceptrons (MLPs). MLPs are excluded for datasets with fewer than 500 samples. “N/A” indicates inapplicability due to sample size constraints (PATE-GAN requires
$n \ge 1000$) or computational limitations (in our experiments, AIM required 8 hours for one synthetic dataset, whereas the other methods completed within seconds or minutes).
Competing methods are formulated under $(\epsilon,\delta)$-DP. To ensure fair comparison, we apply Proposition~\ref{prop:conversions} to convert our $\mu$-GDP guarantee into $(\epsilon,\delta)$-DP. For each dataset, we set $\epsilon = 1$ and $\delta = 1/n^{1.1}$. Predictive performance is measured in terms of relative MSE, averaged across 10 synthetic datasets; see 
Table~\ref{tab:syn_rmse}.

\begin{table}[h]
\centering
\caption{\small Average relative MSE across datasets using different DP-SDG methods. Lowest error per dataset in \textbf{bold}; second lowest \uline{underlined}. All DP methods satisfy $(\epsilon, \delta)$-DP with $\epsilon=1; \delta = 1/n^{1.1}$.}
\label{tab:syn_rmse}
\resizebox{0.5\textwidth}{!}{%
\begin{tabular}{cccccccc}
\toprule
Dataset & Original & AIM & BinAgg & {\footnotesize DP-CTGAN} & {\footnotesize PATE-CTGAN} & {\footnotesize DP-GAN} & {\footnotesize PATE-GAN} \\
\midrule
D1 & 0.286 & 1.400 & \textbf{0.939} & \uline{1.091} & 1.344 & 2.398 & N/A \\
D2 & 1.073 & 1.326 & \textbf{1.015} & 1.126 & \uline{1.054} & 1.984 & N/A \\
D3 & 0.875 & \textbf{1.169} & \uline{2.208} & 17.298 & 4.325 & 12.619 & 8.789 \\
D4 & 0.487 & 0.898 & \textbf{0.731} & 1.289 & \uline{1.001} & 3.522 & 7.775 \\
D5 & 0.033 & 1.058 & \textbf{0.677} & 1.139 & \uline{1.000} & 2.527 & 2.241 \\
D6 & 0.628 & \textbf{0.908} & 1.195 & 1.381 & \uline{1.120} & 2.008 & 2.307 \\
D7 & 0.489 & \uline{0.614} & \textbf{0.584} & 1.231 & 1.039 & 1.488 & 1.753 \\
D8 & 0.683 & \textbf{1.079} & 1.490 & 1.727 & \uline{1.087} & 2.395 & 3.256 \\
D9 & 0.240 & N/A  & \textbf{0.910} & 39.621 & 2.538  & 8.024 & \uline{2.460}  \\
\bottomrule
\end{tabular}%
}
\end{table}

Across these settings, BinAgg achieves the best overall performance: it attains the lowest error on 6 out of 9 datasets and, when not the best, performs very close to the top method. Its performance is followed by AIM and then PATE-CTGAN. 
Beyond accuracy, BinAgg is also markedly faster than the competing methods. AIM requires 221.08 seconds per synthetic dataset on average across datasets D1--D8, whereas BinAgg requires only 0.13 seconds and PATE-CTGAN 6.24 seconds over the same datasets.
This translates to BinAgg being approximately 1,700× faster than AIM,  and 48× faster than PATE-CTGAN.
Details of the running times and computing environments are provided in Appendix~\ref{app:b}. These results highlight that BinAgg delivers both high utility and exceptional efficiency, making it particularly well-suited for regimes where both accuracy and computational efficiency are critical.

\section{Discussion}

We propose BinAgg, a unified DP framework for LR with valid inference and regression-aware SDG. Binning preserves the covariate–response distribution for statistically faithful synthetic data, while aggregation maintains linear relationships necessary for inference in linear models.

Since BinAgg preserves information at the covariate–bin level, it can potentially support regression and inference on arbitrary subsets of covariates without additional privacy cost. Thus, promising future directions include the exploration of model selection procedures such as the F-test, AIC, and BIC when DP uncertainty is properly accounted for. Treatment effect comparisons, a cornerstone of randomized controlled trials, can be carried out directly on synthetic data, offering a path toward privacy-preserving causal inference. Beyond linear models, the released synthetic datasets enable analyses using other statistical or ML methods. 

Our approach is primarily designed for small- to moderate-dimensional settings where binning remains computationally feasible and statistically stable. Extending BinAgg to high-dimensional regimes, where the curse of dimensionality affects partition-based methods, is a natural direction for future work. The method also assumes a prespecified bounded domain for covariates and responses. While such bounds are standard in many applications, developing principled and utility-preserving procedures for privately estimating or adapting bounds is another important direction. In addition, while we establish asymptotic guarantees for estimation and inference, sharper non-asymptotic results would provide a more precise understanding of when the asymptotic normality approximation is appropriate. Finally, improved binning strategies with optimized privacy allocation and adaptive tuning may further enhance accuracy and broaden applicability.

\newpage
\section*{\normalsize Acknowledgments}
This work was supported at Penn State by the Huck Institutes of the Life Sciences through the Dorothy Foehr Huck and J. Lloyd Huck Chair in Data Privacy and Confidentiality, and by a 2025–2026 Rising Researcher Grant from the Institute for Computational and Data Sciences (RRID: SCR\_025154).

\bibliographystyle{agsm}
\bibliography{reference}

\clearpage
\onecolumn

\section*{Appendix}
\addcontentsline{toc}{section}{Appendix}

\setcounter{section}{0}
\renewcommand{\thesection}{\Alph{section}}
\renewcommand{\thesubsection}{\Alph{subsection}}

\subsection{Hyperparameter Detail}
\label{app:a}
\paragraph{Simulation in Section~5.1.}
For hyperparameters of DP-GD, we use grid search over 252 combinations, with learning rate from $\{0.001, 0.005, 0.01, 0.05, 0.1, 0.5, 1\}$, the clipping norm from $\{1, 5, 10, 20, 50, 100\}$, and the number of epochs from $\{1, 5, 10, 20, 50, 100\}$. 
For all other methods, we pre-specify bounds on the covariates and response variable for clipping before data generation: $(0,1)$ for $\bm{x}$ and $(0,2)$, $(0,7)$, $(0,15)$ for $y$ when $d = 1$, $5$, and $10$, respectively.
For BinAgg, the privacy budget is allocated as $\mu_{\text{bin}} : \mu_c : \mu_s : \mu_t = 1 : 3 : 3 : 3$. We use the default value $\theta=0$ in all experiments, noting that negative $\theta$ values lead to more aggressive splitting. 

\paragraph{Simulation in Section~5.2.}
For DP-GD, we use a grid search with learning rates from $\{10^{-6}, 10^{-5}, 10^{-4}, \dots, 1\}$, clipping norms from $\{10^{-6}, 10^{-5}, 10^{-4}, \dots, 10^6\}$, and number of epochs from $\{1, 5, 10, 20, 50\}$, resulting in a total of 455 combinations. This setup is similar to that used in \cite{amin2023easy} for evaluating real datasets. A caveat is that the grid search is conducted on the real data.
For all other methods, we use the non-private data bounds. We acknowledge that these bounds are not DP, but this is acceptable for comparison purposes. In practice, applications are typically conducted by domain experts who have the knowledge to determine appropriate bounds and perform clipping before regression. If tight bounds are difficult to specify, one may either choose conservative bounds or use privatized ones. While conservative bounds often lead to reduced utility, BinAgg demonstrates greater robustness to loose bounds than AdaSSP, as shown in Figure~\ref{fig:loosebounds}.

\subsection{Additional Results for Experiments}
\label{app:b}
\begin{table}[h]
\centering
\caption{\small Datasets used in experiments with size and references.}
\label{tab:datasets}
\resizebox{0.7\textwidth}{!}{%
\begin{tabular}{l l l l}
\toprule
ID & Dataset & Size $(n, d)$ & Reference \\
\midrule
D1 & Intrusion Detection & (182, 4) & \citep{intrusion_detection_in_wsns_715} \\
D2 & Liver Disorders & (345, 6) & \citep{liver_disorders_60} \\
D3 & Auction Verification & (2043, 8) & \citep{auction_verification_713} \\
D4 & Abalone Age & (4177, 10) & \citep{abalone_1} \\
D5 & Parkinson's Telemonitoring & (5875, 21) & \citep{parkinsons_telemonitoring_189} \\
D6 & Wine Quality & (6497, 12) & \citep{wine_quality_186} \\
D7 & Air Quality & (9357, 12) & \citep{air_quality_360} \\
D8 & Appliances Energy & (19735, 27) & \citep{appliances_energy_prediction_374} \\
D9 & Superconductivity & (21263, 81) & \citep{superconductivty_data_464} \\
\bottomrule
\end{tabular}
}
\end{table}

\begin{table}[H]
\centering
\caption{\small Average runtime per synthetic dataset (seconds).}
\label{tab:sdg_summary}
\resizebox{\textwidth}{!}{%
\begin{tabular}{lcccccc}
\toprule
Dataset & AIM & BinAgg & DP-CTGAN & PATE-CTGAN & DP-GAN & PATE-GAN \\
\midrule
Intrusion Detection     & 28.6317 & \textbf{0.0097} & 2.1385 & 1.9045 & 1.3252 & N/A \\
Liver Disorders         & 48.7492 & \textbf{0.0127} & 1.0844 & 1.8050 & 0.7650 & N/A \\
Auction Verification    & 82.9825 & \textbf{0.0282} & 1.8096 & 2.6069 & 0.6688 & 3.0788 \\
Abalone Age             & 130.1068 & \textbf{0.0662} & 5.8339 & 4.7986 & 1.3709 & 4.2810 \\
Parkinson’s Telemonitoring & 407.7839 & \textbf{0.0862} & 13.3774 & 6.6898 & 3.1690 & 6.0159 \\
Wine Quality            & 210.8847 & \textbf{0.0566} & 11.9014 & 10.8458 & 2.0291 & 6.9002 \\
Air Quality             & 236.7886 & \textbf{0.1444} & 24.6687 & 5.2088 & 2.7882 & 5.4930 \\
Appliance Energy        &  622.7202 & \textbf{0.6674} & 113.1582 & 16.0563 & 11.4593 & 12.6062 \\
Superconductivity       &  N/A & \textbf{2.5845} & 968.1989 & 35.1800 & 163.6625 & 24.0455 \\
\bottomrule
\end{tabular}%
}
\end{table}

Computing environment: All experiments were conducted on a laptop with an 8-core AMD Ryzen 7 8845HS CPU and 16 GB of RAM. All synthesizers were run on the same CPU.













\subsection{Algorithm: PrivTree}
\label{app:privtree}
\begin{algorithm}[H]
\caption{\small PrivTree ($D$, $\lambda$, $\theta$, $\tau$) \citep{Zhang2016PrivTreeAD}}
\label{alg:privtree}
\begin{algorithmic}[1]
\State Initialize a tree $\mathcal{T}$ with a root node $v_1$
\State Set $\text{domain}(v_1) = \Omega$, and mark $v_1$ as \texttt{unvisited}
\While{there exists an \texttt{unvisited} node $v$}
    \State Mark $v$ as \texttt{visited}
    \State Compute a biased point count for $v$ with decaying factor $\tau$:
    \State \hskip1em $b(v) = c(v) - \text{depth}(v) \cdot \tau$
    \State Adjust $b(v)$ if it is excessively small:
    \State \hskip1em $b(v) = \max\{b(v), \theta - \tau\}$
    \State Compute a noisy version of $b(v)$: $\hat{b}(v) = b(v) + \text{Lap}(\lambda)$
    \If{$\hat{b}(v) > \theta$}
        \State Split $v$ and add its children to $\mathcal{T}$
        \State Mark the children of $v$ as \texttt{unvisited}
    \EndIf
\EndWhile
\State \Return $\mathcal{T}$ with all point counts removed
\end{algorithmic}
\end{algorithm}

\begin{lemma*}[Corollary 1, \cite{Zhang2016PrivTreeAD}] 
\label{lem:privtreeprivacy}
Let $\kappa$ be the branching factor of tree $\mathcal{T}$.
PrivTree satisfies $\varepsilon$-differential privacy if
\[
\lambda \geq \frac{2\kappa - 1}{\kappa - 1} \cdot \frac{1}{\varepsilon}
\quad \text{and} \quad
\tau = \lambda \cdot \ln \kappa.
\]
\end{lemma*}


\subsection{Proofs}
\label{app:d}
\begin{proof}[\textbf{Proof of Corollary~\ref{thm:equiv}}]
For each bin $k$, let 
\(\tilde{\bm{x}}^{(k,i)}\) denote the privatized version of the $i$-th sample 
within the bin. Given that $\tilde c_k$ is fixed, by construction of the Gaussian mechanism, each privatized 
vector has expectation
\[
\mathbb{E}\!\left(\tilde{\bm{x}}^{(k,i)}\right) = \frac{\bm{s}_k}{\tilde c_k},
\]
and variance
\[
\mathrm{Var}\!\left(\tilde{\bm{x}}^{(k,i)}\right) = 
\frac{\bm{\Delta}_k^{2}}{\tilde c_k\,\mu_s^{2}}.
\]
Because the $\tilde{\bm{x}}^{(k,i)}$’s are independent Gaussian random 
variables, their sum is also Gaussian, with mean equal to the sum of means 
and variance equal to the sum of variances. Consequently,
\[
\tilde{\bm{s}}'_k = \sum_{i=1}^{\tilde c_k} \tilde{\bm{x}}^{(k,i)} 
\sim \mathcal{N}\!\left( \bm{s}_k, \frac{\bm{\Delta}_k^{2}}{\mu_s^{2}} \right).
\]
This distribution matches exactly the law of $\tilde{\bm{s}}_k$, the directly 
privatized statistic. Hence, the two constructions are 
distributionally equivalent. An identical argument, applied to the statistics 
$\tilde{t}'_k$ and $\tilde{t}_k$, shows the same distributional equivalence 
holds for the response terms.
\end{proof}

\bigskip

\begin{proof}[\textbf{Proof of Theorem~\ref{thm:priv}(Algorithm~\ref{alg:regression})}]
The algorithm privatizes the counts $c_k$, the sums of covariates $s_k$, and the sums of responses $t_k$. By Proposition~2.4, the Gaussian mechanism applied to each of these statistics ensures Gaussian DP (GDP) with parameters $\mu_c$, $\mu_s$, and $\mu_t$, respectively.

The initial binning procedure itself incurs a privacy cost 
quantified by $\mu_{\text{bin}}$. By composition of GDP (Proposition~2.2), the overall privacy is given  by $\sqrt{\mu_{\text{bin}}^2 + \mu_s^2 + \mu_t^2 + \mu_c^2}$-GDP. 
By the 
post-processing property (Proposition~2.3), all subsequent steps of the algorithm satisfy the same privacy guarantee.
\end{proof}
\bigskip
\begin{proof}[\textbf{Proof of Theorem~\ref{thm:priv} (Algorithm~\ref{alg:synthetic_data})}]
First, the procedure
privatizes the bin counts $c_k$. By Proposition~2.4, this step satisfies 
$\mu_c$-GDP. 
Next, the algorithm generates $\tilde c_k$ synthetic records per bin $k$. 
Each synthetic feature vector $\tilde{\bm{x}}^{(k,i)}$ is generated by adding 
Gaussian noise calibrated to ensure $\mu_s/\sqrt{\tilde c_k}$-GDP, while each 
synthetic label $\tilde{y}^{(k,i)}$ is privatized with 
$\mu_t/\sqrt{\tilde c_k}$-GDP. Because there are $\tilde c_k$ such records, 
the composition property implies that the total privacy cost across all 
synthetic samples in a given bin is
\[
\sqrt{ \tilde c_k \left( \frac{\mu_s^{2}}{\tilde c_k} + 
                         \frac{\mu_t^{2}}{\tilde c_k} \right) }
= \sqrt{\mu_s^{2} + \mu_t^{2}}.
\]

Combining this contribution with the binning privacy loss and the 
count privatization, and by composition and post-processing, the 
overall privacy guarantee of Algorithm~3 is
\[
\sqrt{\,\mu_{\text{bin}}^{2}+\mu_s^{2}+\mu_t^{2}+\mu_c^{2}\,}\text{-GDP}.
\]
\end{proof}

\bigskip

\begin{proof}[\textbf{Proof of Theorem~\ref{thm:normality}}]
The result concerns the asymptotic distribution of the bias-corrected 
estimator $\tilde{\bm\beta}$. The argument follows from the general theory of 
estimating equations. 

We begin by recalling the noise model. Let 
\(\xi_{\bm s_k}\sim \mathcal{N}(\bm 0, \Sigma_{\bm s_k})\) denote the Gaussian 
noise added to the bin-level summary vector $\bm s_k$, and let 
\(\xi_{t_k}\sim \mathcal{N}(0,\sigma_t^{2})\) denote the noise added to $t_k$. 
Then the privatized quantities are
\[
\tilde{\bm s}_k = \bm s_k + \xi_{\bm s_k}, 
\qquad
\tilde t_k = t_k + \xi_{t_k} = \bm s_k^\top \bm\beta + \eta_k + \xi_{t_k},
\]
where $\eta_k\sim \mathcal{N}(0,\sigma^{2}c_k)$ is the aggregated regression noise.

The estimator $\tilde{\bm\beta}$ is defined implicitly as the solution to the 
estimating equation
\[
\bm Q(\bm b) = \frac{1}{K}\sum_{k=1}^K \bm Q_k(\bm b) = \bm 0,
\]
with
\[
\bm Q_k(\bm b) 
= -\tilde{\bm s}_k \tilde w_k\bigl(\tilde t_k - \tilde{\bm s}_k^\top \bm b\bigr) 
- \tilde w_k D_k \bm b,
\qquad
D_k = \mathbb{E}(\xi_{\bm s_k}\xi_{\bm s_k}^\top).
\]

At the true parameter $\bm\beta$, we have
\begin{align*}
\mathbb{E}[\bm Q_k(\bm\beta)]
&= -\mathbb{E}\Big[(\bm s_k+\xi_{\bm s_k})\tilde w_k
   \bigl(\bm s_k^\top\bm\beta + \eta_k + \xi_{t_k} 
         - (\bm s_k+\xi_{\bm s_k})^\top\bm\beta\bigr)\Big] 
- \mathbb{E}[\tilde w_k D_k \bm\beta] \\
&= \mathbb{E}[\tilde w_k \xi_{\bm s_k}\xi_{\bm s_k}^\top]\bm\beta 
- \mathbb{E}[\tilde w_k] D_k \bm\beta \\
&= 0,
\end{align*}
due to the independence between $\xi_{\bm s_k}$ and $\tilde w_k$.
 Hence, the estimating equation is 
unbiased.

We next analyze the asymptotic distribution. Because $\bm Q(\bm b)$ is linear 
in $\bm b$, its Jacobian with respect to $\bm b$ is constant:
\[
\widetilde M = \frac{\partial \bm Q(\bm b)}{\partial \bm b}
= \frac{1}{K}\sum_{k=1}^K 
  \bigl(\tilde w_k \tilde{\bm s}_k \tilde{\bm s}_k^\top - \tilde w_k D_k\bigr)
= \frac{1}{K}\,\widetilde S^\top \widetilde W \widetilde S - \widetilde D,
\]
where $\widetilde D = \tfrac{1}{K}\sum_{k=1}^K \tilde w_k D_k$. 
A first-order Taylor expansion of $\bm Q(\cdot)$ around $\bm\beta$ yields
\[
\bm 0 = \bm Q(\tilde{\bm\beta}) 
= \bm Q(\bm\beta) + \widetilde M(\tilde{\bm\beta}-\bm\beta),
\]
which rearranges to
\[
\tilde{\bm\beta}-\bm\beta = -\widetilde M^{-1}\bm Q(\bm\beta).
\]

Since the bins $\{B_k\}$ are disjoint, the regression errors $\{\eta_k\}$ are independent,
and the DP noises $\{(\xi_{\bm s_k},\xi_{t_k})\}$ are generated independently across bins,
the vectors $\{\bm Q_k(\bm\beta)\}_{k=1}^K$ are independent
with finite second moments, so the multivariate CLT applies.
 We then have
\[
\sqrt{K}\,\bm Q(\bm\beta)
=\frac{1}{\sqrt K}\sum_{k=1}^K \bm Q_k(\bm\beta)
\ \xrightarrow{d}\ \mathcal N(\bm 0, H),
\qquad
H=\lim_{K\to\infty}\frac{1}{K}\sum_{k=1}^K \operatorname{Var}\!\big(\bm Q_k(\bm\beta)\big).
\]
If $\widetilde M \xrightarrow{p} M$ with $M$ nonsingular, Slutsky's theorem yields
\[
\sqrt{K}\,(\tilde{\bm\beta}-\bm\beta)
\ \xrightarrow{d}\ \mathcal N\!\big(\bm 0,\; M^{-1}HM^{-1}\big).
\]

A consistent estimator of $H$ is given by its corrected sample analogue
\[
\widetilde H=\frac{1}{K-d}\sum_{k=1}^K \widetilde{\bm Q}_k\widetilde{\bm Q}_k^\top, 
\quad 
\widetilde{\bm Q}_k 
= \tilde{\bm s}_k \tilde w_k(\tilde t_k-\tilde{\bm s}_k^\top \tilde{\bm\beta})
  + \tilde w_k D_k \tilde{\bm\beta},
\]
so a consistent estimator of $\operatorname{Var}(\tilde{\bm\beta})$ is
\[
\widetilde\Sigma=\frac{1}{K}\,\widetilde M^{-1}\widetilde H\,\widetilde M^{-1}.
\]
This establishes asymptotic normality of the bias-corrected estimator, along with consistent covariance estimation.
\end{proof}

\end{document}